\documentclass[10pt,twocolumn,letterpaper]{article}

\usepackage{cvpr}
\usepackage{times}
\usepackage{epsfig}
\usepackage{graphicx}
\usepackage{amsmath}
\usepackage{amssymb}
\usepackage{amsthm}
\usepackage{csquotes}
\graphicspath{ {./images/} }

\usepackage[breaklinks=true,bookmarks=false]{hyperref}

\hypersetup{
    colorlinks=true,
    filecolor=magenta,      
    urlcolor=magenta,
}

\cvprfinalcopy 

\def\cvprPaperID{****} 

\DeclareMathOperator*{\minimize}{argmin}

\begin{document}

\title{Unsupervised Domain Adaptation using Feature-Whitening  and  Consensus Loss}

\author{Subhankar Roy\textsuperscript{1,2}, Aliaksandr Siarohin\textsuperscript{1}, Enver Sangineto\textsuperscript{1}, Samuel Rota Bul\`{o}\textsuperscript{3},\\ Nicu Sebe\textsuperscript{1} and Elisa Ricci\textsuperscript{1,2} \\
\textsuperscript{1}DISI, University of Trento, Italy, \textsuperscript{2}Fondazione Bruno Kessler, Trento, Italy, \textsuperscript{3}Mapillary Research  \\
{\tt\small \{subhankar.roy, aliaksandr.siarohin, enver.sangineto, niculae.sebe, e.ricci\}@unitn.it}
\vspace{-3mm}\and
{\tt\small samuel@mapillary.com}
}

\maketitle

\begin{abstract}
A classifier trained on a dataset seldom works on other datasets obtained under different conditions due to domain shift. This problem is commonly addressed by domain adaptation methods. In this work we introduce a novel deep learning framework which unifies different paradigms in unsupervised domain adaptation. Specifically, we propose domain alignment layers which implement feature whitening for the purpose of matching source and target feature distributions. Additionally, we leverage the unlabeled target data by proposing the Min-Entropy Consensus loss, which  regularizes  training while avoiding the adoption of many user-defined hyper-parameters. We report results on publicly available datasets, considering both digit classification and object recognition tasks. We show that, in most of our experiments, our approach improves upon previous methods, setting new state-of-the-art performances.
\end{abstract}

\section{Introduction}
\label{sec:introduction}

Deep learning methods have been  successfully applied to different visual recognition tasks, 
demonstrating an excellent generalization ability. However, analogously to other statistical machine learning techniques, deep neural networks also suffer from the problem of {\em domain shift} \cite{torralba2011unbiased},
which is observed when 
predictors trained on a dataset do not perform well when applied to novel domains. 

Since collecting annotated training data from every possible domain is expensive and sometimes even impossible, over the years several Domain Adaptation (DA) methods \cite{pan2010survey,csurka2017domain} have been proposed. DA approaches leverage labeled data in a source domain in order to learn an accurate prediction model for a target domain.
Specifically, in the special case of 
Unsupervised Domain Adaptation (UDA),  
no annotated target data are available at training time. Note that, even if target-sample labels are not available, unlabeled data can and usually are exploited at training time.


\begin{figure*}[!h]
    \centering
    \includegraphics[width=0.9\textwidth]{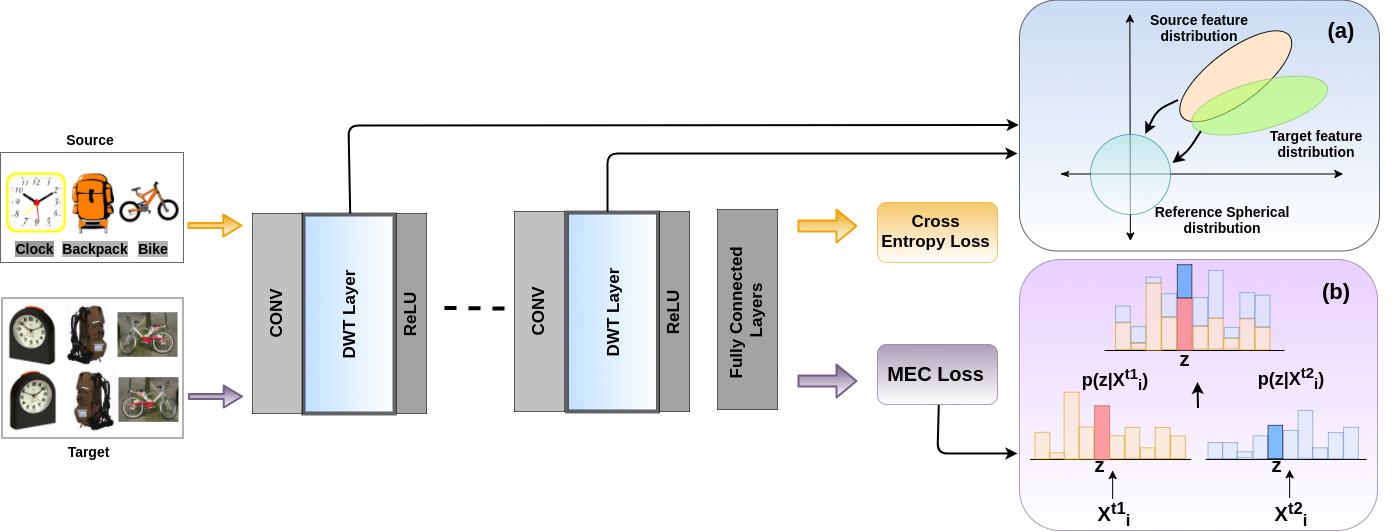}
    \caption{Overview of the proposed deep architecture embedding our DWT layers and trained with the proposed MEC loss. (a) Due to domain shift the source and the target data have different marginal feature distributions. Our DWT estimates these distributions using dedicated sample batches and then ``whitens'' them 
    projecting them into a common, spherical distribution. (b) The proposed MEC loss univocally selects a pseudo-label $z$ that maximizes the agreement between two perturbed versions $\mathbf{x}_i^{t_1}$ and $\mathbf{x}_i^{t_2}$ of the same target sample. 
    }
    \label{fig:overall_pipenline}
\end{figure*}

Most UDA methods attempt to reduce the domain shift by directly aligning the source and target marginal distributions.
Notably, approaches based on the {\em Correlation Alignment} paradigm model domain data distributions in terms of 
their second-order statistics. 
Specifically, they match 
distributions by minimizing a loss function which corresponds to the difference between the source and the target covariance matrices obtained using the network's last-layer activations \cite{sun2016return, sun2016deep,morerio2017minimal}.
Another recent and successful UDA paradigm exploits {\em domain-specific alignment layers}, derived from Batch Normalization (BN) \cite{ioffe2015batch}, which are directly embedded within the deep network 
\cite{carlucci2017autodial,li2016revisiting,mancini2018boosting}.
Other prominent research directions in UDA correspond to those methods which also 
exploit the target data posterior distribution.
For instance, the {\em entropy minimization} paradigm adopted in \cite{ carlucci2017autodial,saito2017asymmetric, haeusser2017associative},
enforces  the network's prediction probability distribution on each target sample to be peaked with respect to some (unknown) class, thus penalizing high-entropy target predictions.
On the other hand, the {\em consistency-enforcing} paradigm \cite{sajjadi2016regularization,french2018self,tarvainen2017mean} is based on specific loss functions which penalize inconsistent predictions over perturbed copies of the same target samples. 

In this paper we propose to unify the above  paradigms by introducing two main novelties. First, we  align  the  source and the target data distributions using covariance matrices similarly to \cite{sun2016return, sun2016deep, morerio2017minimal}. However, instead of using a loss function computed on the last-layer activations, 
we use domain-specific alignment layers which compute domain-specific  covariance matrices of intermediate features. These layers  \enquote{whiten} the source and the target features and   project  them into a common spherical distribution (see   Fig.~\ref{fig:overall_pipenline} (a), blue box). 
We call this alignment strategy {\em Domain-specific Whitening Transform} (DWT). Notably, our approach generalizes previous BN-based DA methods  \cite{carlucci2017autodial,li2016revisiting,mancini2018kitting} which do not consider inter-feature correlations and rely only on feature standardization.



The second novelty we introduce is a novel loss function, the Min-Entropy Consensus (MEC) loss, which merges both the  entropy  \cite{carlucci2017autodial,saito2017asymmetric, haeusser2017associative} and the consistency \cite{french2018self} loss function.
The motivation behind our proposal is to avoid the tuning of the many hyper-parameters which are  typically required when considering several loss terms and, specifically, the confidence-threshold hyper-parameters \cite{french2018self}. 
Indeed, due to the mismatch between the source and the target domain, and because of the unlabeled  target-data assumption, 
hyper-parameters are hard to be tuned in UDA 
\cite{morerio2017minimal}. 
The proposed MEC loss simultaneously encourages coherent predictions between two perturbed versions of the same target sample and exploits these predictions as pseudo-labels for  training. 
(Fig.~\ref{fig:overall_pipenline} (b), purple box). 

We plug our proposed DWT and the MEC loss into different network architectures and we empirically show a significant boost in performance. In particular, we achieve state-of-the-art results in different UDA benchmarks: {MNIST} \cite{lecun1998gradient},
USPS \cite{friedman2001elements}, 
SVHN \cite{netzer2011reading}, CIFAR-10, STL10 \cite{coates2011analysis} and Office-Home \cite{venkateswara2017deep}. Our code\footnote{Code available at \url{https://github.com/roysubhankar/dwt-domain-adaptation}} is publicly available. 
\section{Related Work}
\label{sec:rel_works}

\textbf{Unsupervised Domain Adaptation.} Several previous works have addressed the problem of DA, considering both shallow models and deep architectures. In this section we focus on only deep learning methods for UDA, as these are the closest to our proposal.

UDA methods mostly differ in the strategy used to reduce the discrepancy between the source and the target feature distributions and can be grouped in different categories.
The first category includes methods modeling the domain distributions in terms of their first and second order statistics. For instance, some works aim at reducing the domain shift by minimizing the Maximum Mean Discrepancy \cite{long2015learning,long2016deep,venkateswara2017deep} and describe distributions in terms of their first order statistics. Other works consider also second-order statistics using the {\em correlation alignment} paradigm (Sec.~\ref{sec:introduction}) \cite{sun2016deep,morerio2017minimal}.
Instead of introducing additional loss functions, more recent works deal with the domain-shift problem by directly embedding into a deep network  {\em domain alignment layers} which exploit BN~\cite{li2016revisiting,carlucci2017autodial,mancini2018boosting, mancini2019adagraph}.

A second category of methods include approaches which learn  domain-invariant deep representations. For instance, in \cite{ganin2014unsupervised}  a gradient reversal layer learns  discriminative domain-agnostic representations. Similarly, in \cite{tzeng2015simultaneous} a domain-confusion loss is introduced, encouraging the network to learn features robust to the domain shift. Haeusser \textit{et al.} \cite{haeusser17} present Associative Domain Adaptation, an approach which also learns domain-invariant embeddings.

A third category includes methods based on Generative Adversarial Networks (GANs) \cite{russo17sbadagan,Bousmalis:Google:CVPR17, Taigman2016UnsupervisedCI,Shrivastava:arXiv:16, sankaranarayanan2017generate}. The main idea behind these approaches is to directly transform images from the target domain to the source domain. 
While GAN-based methods are especially successful in adaptation from synthetic to real images and in case of non-complex datasets, they have limited capabilities for complex images.


{\em Entropy minimization}, first introduced in \cite{grandvalet2004semisupervised}, is a common strategy in semi-supervised learning \cite{zhu2005semi}. In a nutshell, it consists in exploiting the high-confidence predictions of unlabeled samples as pseudo-labels. Due to its effectiveness, several popular UDA methods \cite{russo17sbadagan,carlucci2017autodial,saito2017asymmetric,long2016deep} have adopted the entropy-loss for training deep networks. 

Another popular paradigm in UDA, which we refer to as the {\em consistency-enforcing} paradigm, is realized by perturbing the target samples and then imposing some consistency between the predictions of two perturbed versions of the same target input. Consistency is imposed by defining appropriate loss functions, as shown in \cite{saito2017asymmetric, french2018self,sajjadi2016regularization}. The consistency loss paradigm is effective but it becomes uninformative if the network produces uniform probability distributions for corresponding target samples. Thus, previous methods also integrate a Confidence Thresholding (CT) technique \cite{french2018self}, in order to discard unreliable predictions. Unfortunately, CT introduces additional user-defined and dataset-specific hyper-parameters which are  difficult to tune in an UDA scenario \cite{morerio2017minimal}. 
Differently, as demonstrated in our experiments, our  MEC loss eliminates the need of CT and the corresponding hyper-parameters.

\textbf{Feature Decorrelation.} Recently, Huang \textit{et al.} \cite{huang2018decorrelated}
and Siarohin \textit{et al.} \cite{2018arXiv180600420S} 
proposed to replace BN with feature 
whitening in a discriminative and generative setting, respectively.
However, none of these works consider a DA problem.  
We show in this paper that feature whitening can be used 
to align the source and the target marginal distributions using layer-specific covariance matrices without the need of a dedicated loss function as in previous correlation alignment methods.

\vspace{-1mm}
\section{Method}
In this section we present the proposed UDA approach. Specifically, after introducing some preliminaries, we describe
our Domain-Specific Whitening Transform and, finally, the
proposed Min-Entropy Consensus loss.

\subsection{Preliminaries}
\label{sec:Preliminaries}
Let ${\cal S} = \{ (I_j^s, y_j^s) \}_{j=1}^{n_s}$ be the labeled source dataset, where $I_j^s$ is an image and
 $y_j^s \in {\cal Y} = \{1, 2 \dots, C 
 \}$ its associated label,
  and 
${\cal T} = \{ I_i^t \}_{i=1}^{n_t}$ be the unlabeled target dataset. The goal of UDA is to learn a predictor for the target domain by using samples from both ${\cal S}$ and ${\cal T}$. Learning a predictor for the target domain is not trivial because of the issues discussed in Sec. \ref{sec:introduction}.


A common technique to reduce domain shift is to use BN-based layers inside a network, such as to project the source and target feature distributions to a reference distribution through feature standarization. As mentioned in Sec. \ref{sec:introduction}, in this work we propose to replace feature standardization with whitening, where the whitening operation is domain-specific. Before introducing the proposed whitening-based distribution alignment, we recap below BN.
Let $B = \{ \mathbf{x}_1, ..., \mathbf{x}_m \}$ be a mini-batch of 
$m$ input samples to a given network layer, where each element  $\mathbf{x}_i \in B$ is a $d$-dimensional feature vector, \ie $\mathbf{x}_i \in \mathbb{R}^d$.
Given $B$, in BN each $\mathbf{x}_i \in B$ is transformed as follows:

\begin{equation}
\label{eq.BN}
\textit{BN}(x_{i,k}) 
= \gamma_k \frac{x_{i,k} - \mu_{B,k}}{\sqrt{\sigma_{B,k}^2 + \epsilon}} + \beta_k,
\end{equation}

\noindent
where $k$ ($1 \leq k \leq d$) indicates the $k$-th dimension of the data,  $\mu_{B,k}$ and 
$\sigma_{B,k}$ are, respectively, the mean and the standard deviation computed with respect to the $k$-th dimension of the samples in $B$ and $\epsilon$ is a constant used to prevent numerical instability. Finally, $\gamma_k$ and $ \beta_k$ are scaling and shifting learnable parameters.

In the next section we present our DWT, while in Sec.~\ref{sec:Consensus} we present the proposed MEC loss. It is worth noting that each proposed component can be plugged independently in a network without having to rely on each other.

\subsection{Domain-specific Whitening Transform}
\label{sec:Whitening}

As stated above, BN is based on a per-dimension {\em standardization} of each sample $\mathbf{x}_i \in B$. Hence, once normalized, the batch samples may still have correlated feature values. Since our goal is to use feature normalization in order to alleviate the domain-shift problem (see below), we argue
that plain standardization is not enough to align the source and the target marginal distributions. For this reason we propose to use 
Batch Whitening (BW) instead of BN, which is defined as:
\vspace{-2mm}
\begin{align}
\textit{BW}(x_{i,k}; \Omega) 
&= \gamma_k \hat{x}_{i,k}   + \beta_k, \label{eq.scale-shift-after-whitening} \\
\hat{\mathbf{x}}_i &= W_B (\mathbf{x}_i - \boldsymbol{\mu}_B). \label{eq.WC-whitening}
\end{align}

In Eq.~\eqref{eq.WC-whitening}, the vector $\boldsymbol{\mu}_B$ is the mean of the elements in $B$ (being $\mu_{B,k}$ its $k$-th component) while the matrix $W_B$ is such that: $W_B^\top W_B = \Sigma_B^{-1}$, where $\Sigma_B$ is the covariance matrix computed using $B$. 
$\Omega = (\boldsymbol{\mu}_B, \Sigma_B)$ are the batch-dependent first and  second-order statistics. Eq.~\eqref{eq.WC-whitening} performs the {\em whitening} of $\mathbf{x}_i$ and the resulting set of vectors $\hat{B} = \{ \hat{\mathbf{x}}_1, ..., \hat{\mathbf{x}}_m \}$ lie in a spherical distribution (i.e., with a covariance matrix equal to the identity matrix).

Our network takes as input two different batches of data, randomly extracted from ${\cal S}$ and ${\cal T}$, respectively.
Specifically, given any arbitrary layer $l$ in the network, let $B^s = \{ \mathbf{x}_1^s, ..., \mathbf{x}_m^s \}$ and
$B^t = \{ \mathbf{x}_1^{t}, ..., \mathbf{x}_m^{t} \}$ denote the batch of intermediate output activations, from  layer $l$, for the source and target domain, respectively. Using Eq.~\eqref{eq.scale-shift-after-whitening}-\eqref{eq.WC-whitening} we can now define our Domain-specific Whitening Transform 
(DWT). Let $x^s$ and $x^t$ denote the inputs to the DWT layer from the source and the target domain, respectively. Our  DWT is defined as follows (we drop the sample index $i$ and dimension index $k$ for the sake of clarity):

\vspace{-5mm}
\begin{align}
    \textit{DWT}(x^{s}; \Omega^s) 
    &= BW(x^{s}, \Omega^{s}),\\
\textit{DWT}(x^{t}; \Omega^t) 
    &= BW(x^{t}, \Omega^{t}).
\end{align}

We  estimate separate statistics ($\Omega^s = (\boldsymbol{\mu}^{s}_B, \Sigma^{s}_{B})$ and $\Omega^t = (\boldsymbol{\mu}^{t}_B, \Sigma^{t}_{B})$) for $B^s$ and $B^t$ and use them for whitening the corresponding activations, projecting the two batches into a common spherical distribution (Fig.~\ref{fig:overall_pipenline} (a)). 

 $W^{s}_{B}$ and $W^{t}_{B}$ are computed following the approach described in \cite{2018arXiv180600420S}, 
which is based on the Cholesky decomposition \cite{Cholesky}. The latter is faster \cite{2018arXiv180600420S} than the ZCA-based whitening \cite{ZCA-whitening} adopted in \cite{huang2018decorrelated}.
In the Supplementary Material we provide more details on how $W^{s}_{B}$ and $W^{t}_{B}$ are computed.
Differently from \cite{2018arXiv180600420S} we replace the \enquote{coloring} step after whitening with simple scale and shift 
operations, thereby preventing the introduction of extra parameters in the network. 
Also, differently from \cite{2018arXiv180600420S} we use {\em feature grouping} \cite{huang2018decorrelated} (Sec.~\ref{sec:implementation_details}) in order to make the batch-statistics estimate more robust when $m$ is small and $d$ is large.
During training, the DWT layers accumulate the statistics for the target domain using
a moving average of the batch statistics ($\Omega^{t}_{avg}$).

In summary, the proposed DWT layers replace the  correlation alignment of the last-layer feature activations with the intermediate-layer feature whitening, performed at different levels of abstraction. In Sec. \ref{sec:implementation_details} we show that BN-based domain alignment layers \cite{li2016revisiting,carlucci2017autodial} can be seen as a special case of DWT layers.

\vspace{-2mm}
\subsubsection{Implementation Details}
\label{sec:implementation_details}

Given a typical block (Conv layer $\rightarrow$ BN $\rightarrow$ ReLU) of a CNN, we replace the BN layer with our proposed DWT layer (see in Fig. \ref{fig:overall_pipenline}), obtaining: (Conv layer $\rightarrow$ DWT $\rightarrow$ ReLU).
Ideally, in order to project the source and target feature distributions to a reference one, the DWT layers should perform full-feature whitening using a $d$ $\times$ $d$ whitening matrix, where $d$ is the number of features. However, the computed covariance matrix $\Sigma_{B}$ can be ill-conditioned if $d$ is large and $m$ is small. 
For this reason,
unlike \cite{2018arXiv180600420S} and similar to \cite{huang2018decorrelated} we use {\em feature grouping}, where the features are grouped into subsets of size $g$. This results in better-conditioned 
covariance matrices but into partially whitened features. In this way we reach a compromise between full-feature whitening and numerical stability.
Interestingly, when $g=1$, the whitening matrices reduce to diagonal matrices, thus realizing feature standardization as in \cite{carlucci2017autodial,li2016revisiting}.

\subsection{Min-Entropy Consensus Loss}
\label{sec:Consensus}
The impossibility of using the cross-entropy loss on the unlabeled target samples is commonly circumvented using some common unsupervised loss, such as the entropy  \cite{carlucci2017autodial,saito2017asymmetric} or the consistency loss \cite{french2018self, sajjadi2016regularization}. While minimizing the entropy loss ensures that the predictor maximally separates the target data, minimization of the consistency loss forces the predictor to deliver consistent predictions for target samples coming from identical (yet unknown) category. Therefore, given the importance of exploiting better the unlabeled target data and the limitations of the above two losses (see Sec. \ref{sec:introduction}), we propose a novel Min-Entropy Consensus (MEC) loss within the framework of UDA. We  explain below how MEC loss merges both the entropy and the consistency loss into a single unified function.

Similar to the consistency loss, the proposed MEC loss requires input data perturbations. Unless otherwise explicitly specified, we apply common data-perturbation techniques on both ${\cal S}$ and ${\cal T}$ using affine transformations and Gaussian blurring operations.
When we use the MEC loss, the network is fed with three batches instead of two. Specifically, apart from $B^{s}$, we use two different target batches ($B^t_1$ and $B^t_2$), which contain duplicate pairs of images differing only with respect to the adopted image perturbation.

 Conceptually, we can think of this pipeline as three different networks with three separate domain-specific statistics $\Omega^{s}$, $\Omega^{t}_{1}$ and $\Omega^{t}_{2}$ but with shared network weights. However, 
 since both $B^t_1$ and $B^t_2$ are drawn from the same distribution,
 we estimate a single $\Omega^{t}$ using both the target batches ($B^{t}_{1} \bigcup B^{t}_{2}$).
 As an additional advantage, this makes it possible to use $2m$ samples for computing  $\Sigma^{t}_{B}$.
 
 Let  $B^s = \{ \mathbf{x}_1^s, ..., \mathbf{x}_m^s \}$,
$B^t_1 = \{ \mathbf{x}_1^{t_1}, ..., \mathbf{x}_m^{t_1} \}$ and
$B^t_2 = \{ \mathbf{x}_1^{t_2}, ..., \mathbf{x}_m^{t_2} \}$ be three batches of the last-layer activations. 
Since the source samples are labeled, the cross-entropy loss ($L^s$) can be used in case of $B^s$:

\vspace{-3mm}
\begin{equation}
\label{eq.source-cross-entropy-loss}
L^s(B^s) = - \frac{1}{m} \sum_{i=1}^m \log p(y_i^s | \mathbf{x}_i^s),
\end{equation}

\noindent
where $p(y_i^s | \mathbf{x}_i^s)$ is the (soft-max-based) probability prediction  assigned by the network to a sample $\mathbf{x}_i^s \in B^s$  with respect to its ground-truth label $y_i^s$.
However, ground-truth labels are not available for target samples. For this reason, we propose the following MEC loss ($L^t$):
\vspace{-2mm}
\begin{equation}
L^t(B^t_1, B^t_2) = \frac{1}{m} \sum_{i=1}^m  \ell^t(\mathbf x^{t_1}_i,\mathbf x_i^{t_2}), \label{eq.consensus-entropy}
\end{equation}
\vspace{-3mm}
\begin{equation}
\ell^t(\mathbf x^{t_1}_i,\mathbf x_i^{t_2})=  -\frac{1}{2}\max_{y \in {\cal Y}}  \Big (\log p(y | \mathbf{x}_i^{t_1}) + \log p(y | \mathbf{x}_i^{t_2}) \Big). \label{eq.pointwise_loss} 
\end{equation}

\noindent
In Eq.~\eqref{eq.pointwise_loss}, $\mathbf{x}_i^{t_1} \in B^t_1$ and $\mathbf{x}_i^{t_2} \in B^t_2$ are  activations of two corresponding perturbed target samples. 

The intuitive idea behind our proposal is that, similarly to consistency-based losses \cite{french2018self,sajjadi2016regularization}, 
since $\mathbf{x}_i^{t_1}$ and $\mathbf{x}_i^{t_2}$ correspond to the same image, the network should provide similar predictions. However, unlike the aforementioned  methods which compute the L2-norm or the binary cross-entropy between these predictions, the proposed MEC loss finds the class $z$ such that $z = \minimize_{y \in {\cal Y}}  \Big(\log p(y | \mathbf{x}_i^{t_1}) + \log p(y | \mathbf{x}_i^{t_2})\Big)$. $z$ is the class in which the posteriors corresponding to $\mathbf{x}_i^{t_1}$ and $\mathbf{x}_i^{t_2}$ maximally agree. We then use $z$  as the 
pseudo-label, which can be selected without ad-hoc confidence thresholds.
In other words, instead of using high-confidence thresholds to discard unreliable target samples 
\cite{french2018self}, 
we use all the samples but we  backpropagate the error with respect to only  $z$.

The dynamics of MEC loss is the following. 
First, similarly to the consistency losses, it forces the network to provide coherent predictions.
 Second, differently from consistency  losses, which are prone to attain a near zero value with uniform  posterior  distributions, 
 it enforces peaked predictions.
 See the Supplementary Material for a more formal relation
between the MEC loss and both entropy and consistency loss.

The final loss $L$ is a weighted
sum of $L^s$ and $L^t$: $L(B^s, B^t_1, B^t_2) = L^s(B^s) + \lambda L^t(B^t_1, B^t_2)$.

\subsection{Discussion}
\label{sec:Discussion}

The proposed DWT  generalizes the BN-based DA approaches by decorrelating the batch features. 
Besides the analogy   with the 
correlation-alignment methods mentioned in Sec.~\ref{sec:introduction}, in which covariance matrices are used to estimate and align the source and the target distributions, 
a second reason for which we believe that full-whitening is important is due to the relation  between feature normalization and the smoothness of the loss 
\cite{shu2018dirt,2018arXiv180510694K,huang2018decorrelated,A-smooth,B-smooth}.
For instance, previous works \cite{A-smooth,B-smooth} showed that better conditioning of the input-feature covariance matrix leads to better conditioning of the Hessian of the loss function, making the gradient descent weight updates closer to Newton updates. 
However, BN only performs standardization, which
barely improves the conditioning of the covariance matrix when the features are correlated \cite{huang2018decorrelated}. 
Conversely, feature whitening completely decorrelates the batch samples, thus potentially improving the smoothness of the landscape of the loss function. 

The importance of a smoothed loss function is even higher when entropy-like losses on unlabeled data are used. For instance, Shu \etal \cite{shu2018dirt} 
showed that minimizing the  entropy forces the classifier to be confident on the unlabeled
target data, thus potentially driving the classifier’s decision boundaries away from the target data. 
However, without a locally-Lipschitz constraint on the loss function (\ie with a non smoothed loss landscape),
the decision boundaries can be placed close to the training samples even when the   entropy is
minimized \cite{shu2018dirt}. 
Since our MEC loss is related with both the entropy and the consistency loss, we employ DWT also  to improve the smoothness of our loss function in order to alleviate overfitting phenomena related to the use of unlabeled data.
\vspace{-2mm}

\section{Experiments}
\label{sec:exeriments}

In this section we provide details about our implementation and training protocols and we report our experimental evaluation. We conduct experiments on both  small and large-scale datasets and we compare our method with state-of-the-art approaches. We also present an ablation study to analyze the impact of each of our contributions on the classification accuracy. 

\subsection{Datasets}
\begin{figure}[t]
    \centering
    \begin{tabular}{cc}
         \includegraphics[width=100px]{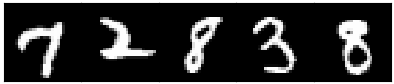} & \includegraphics[width=100px]{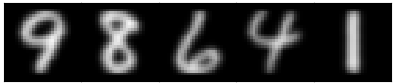}\\
         \multicolumn{2}{c}{(a) MNIST $\leftrightarrow$ USPS}\\
         \includegraphics[width=100px]{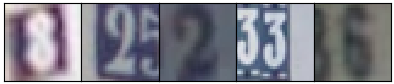} & \includegraphics[width=100px]{{mnist.png}}\\
         \multicolumn{2}{c}{(b) SVHN $\leftrightarrow$ MNIST}\\
         \includegraphics[width=100px]{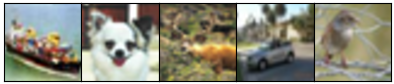} & \includegraphics[width=100px]{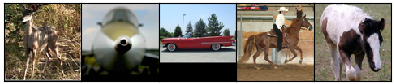}\\
         \multicolumn{2}{c}{(c) CIFAR-10 $\leftrightarrow$ STL}\\
    \end{tabular}
    \caption{Small image datasets used in our experiments.}
    \label{fig:small_image_data_set}
    \vspace{-0.4cm}
\end{figure}

We conduct experiments on the following datasets:

\textbf{MNIST $\leftrightarrow$ USPS.} 
The \textbf{MNIST} dataset \cite{lecun1998gradient} contains grayscale images (28 $\times$ 28 pixels) depicting handwritten digits ranging from 0 to 9. The \textbf{USPS} \cite{friedman2001elements} dataset is similar to MNIST, but images have smaller resolution (16 $\times$ 16 pixels). See Fig.~\ref{fig:small_image_data_set}(a) for sample images.

\textbf{MNIST $\leftrightarrow$ SVHN.} Street View House Number (SVHN) \cite{netzer2011reading} images are 32 $\times$ 32 pixels RGB images. Similarly to the MNIST dataset digits range from 0 to 9. However, in SVHN images have variable colour intensities and depict non-centered digits. Thus, there is a significant domain shift with respect to MNIST (Fig.~\ref{fig:small_image_data_set}(b))

\textbf{CIFAR-10 $\leftrightarrow$ STL}: CIFAR-10 is a 10 class dataset of RGB images depicting generic objects and with resolution 32 $\times$ 32 pixels.
STL \cite{coates2011analysis} is similar to the CIFAR-10, except it has fewer labelled training images per class and has images of resolution 96 $\times$ 96 pixels. The non-overlapping classes - \enquote{frog} and \enquote{monkey} are removed from CIFAR-10 and STL, respectively. Samples are shown in Fig.~\ref{fig:small_image_data_set}.(c).

\begin{figure}[t]
    \centering
    \includegraphics[width=1\columnwidth]{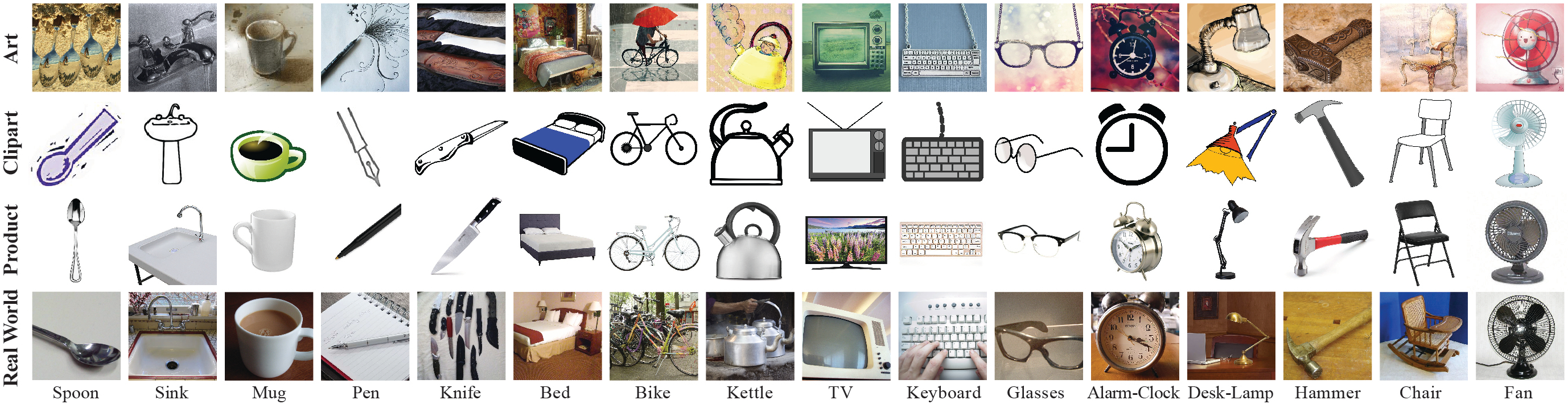}
    \caption{Sample images from the Office-Home dataset.}
    \label{fig:office_home_dataset}
      \vspace{-0.4cm}
\end{figure}

\textbf{Office-Home}: The Office-Home \cite{venkateswara2017deep} dataset comprises 4 distinct domains, each corresponding to 65 different categories (Fig.~\ref{fig:office_home_dataset}). There are 15,500 images in the dataset, thus this represents large-scale benchmark for testing domain adaptation methods. The domains are: \texttt{Art}(\textbf{Ar}), \texttt{Clipart} (\textbf{Cl}), \texttt{Product} (\textbf{Pr}) and \texttt{Real World} (\textbf{Rw}).


\subsection{Experimental Setup}
\label{sec:net_train}

To fairly compare our method with other UDA approaches, in the digits experiments we adopt the same base networks proposed in \cite{ganin2016domain}. 
For the CIFAR-10$\leftrightarrow$STL experiments we use the network described in \cite{french2018self}.
We train the networks using the Adam optimizer \cite{kingma2014adam} with a mini-batch of cardinality $m$ = 64 samples, an initial learning rate of 0.001 and weight decay of 5 $\times$ $10^{-4}$. The networks are trained for a total of 120 epochs with learning rate being decreased by a factor of 10 after 50 and 90 epochs.
We use the SVHN $\rightarrow$ MNIST setting to fix the value of the hyperparameter $\lambda$ to 0.1 and to set group size ($g$) equal to 4. These hyperparameters values are used for all the datasets. 

In the Office-Home dataset experiments we use a ResNet-50 \cite{he2016deep} following \cite{long2017conditional}. 
In our experiments we modify ResNet-50 by replacing the first BN layer and the BN layers in the first residual block (with 64 features) with DWT layers. The network is initialized with weights taken from a pre-trained model trained on the ILSVRC-2012 dataset. We discard the final fully-connected layer and we replace it with a randomly initialized fully-connected  layer with 65 output logits. During training, each domain-specific batch is limited to $m$ = 20 samples (due to GPU memory constraints). The SGD optimizer is used with an initial learning rate of $10^{-2}$ for the randomly initialized final layer and $10^{-3}$ for the rest of the trainable parameters of the network. The network is trained for a total of 60 epochs where one \enquote{epoch} is the pass through the entire data set having the lower number of training samples. The learning rates are then decayed by a factor of 10 after 54 epochs. Differently from the small-scale datasets experiments, where target samples have predefined train and test splits, in the Office-Home experiments, all the target samples (without labels) are used during training and evaluation. 

To demonstrate the effect our contributions, we consider three different variants for the proposed method. In the first variant (denoted as \textbf{DWT} in Sec.~\ref{sec:Whitening}), we only consider DWT layers {\em without} the proposed MEC loss. In practice, in the considered network architectures we replace the BN layers which follows the convolutional layers with DWT layers. Supervised cross-entropy loss is used for the labeled source samples and the entropy-loss as in \cite{carlucci2017autodial} is used for the unlabeled target samples. No data-augmentation is used here.
In the second variant, denoted as \textbf{DWT-MEC}, we also exploit the proposed MEC loss (this corresponds to our full method). In this case we need perturbations of the input data, which are obtained by following some basic data-perturbation schemes like image translation by a factor of [0.05, 0.05], Gaussian blur ($\sigma = 0.1$) and random affine transformation as proposed in \cite{french2018self}. In the third variant (\textbf{DWT-MEC (MT)}) we  plug our proposed DWT layers and the MEC loss in the Mean-Teacher (MT) training paradigm  \cite{tarvainen2017mean}.

\subsection{Results}
In this section we present an extensive experimental analysis of our approach, showing both the results of an ablation study and a comparison with state-of-the-art methods.

\begin{table*}[t]
    \centering
	\setlength{\tabcolsep}{5pt}
    \begin{tabular}{l|cccccc}
        \hline
        \hline
         Methods & \shortstack{Source\\Target} & \shortstack{MNIST\\USPS} & \shortstack{USPS\\MNIST}  & \shortstack{SVHN\\MNIST} & \shortstack{MNIST\\SVHN} \\ 
         \hline
         \hline
         Source Only & & 78.9 &  57.1$\pm$1.7 & 60.1$\pm$1.1 &  20.23$\pm$1.8\\
         \hline
         w/o augmentation\\
         CORAL \cite{sun2016return} & &  81.7 & - & 63.1 & -\\
         MMD \cite{tzeng2015simultaneous} & &  81.1 & - & 71.1 & -\\
         DANN \cite{ganin2016domain} & & 85.1 & 73.0$\pm$2.0 & 73.9 & 35.7\\
         DSN \cite{bousmalis2016domain} & & 91.3 & - & 82.7 & -\\
         CoGAN \cite{liu2016coupled} & & 91.2 & 89.1$\pm$0.8 & - & -\\
         ADDA \cite{tzeng2017adversarial} & & 89.4$\pm$0.2 & 90.1$\pm$0.8 & 76.0$\pm$1.8 & -\\
         DRCN \cite{ghifary2016deep} & & 91.8$\pm$0.1 & 73.7$\pm$0.1 & 82.0$\pm$0.2 & 40.1$\pm$0.1\\
         ATT \cite{saito2017asymmetric} & & - & - & 86.20 & 52.8\\
         ADA \cite{haeusser2017associative} & & - & - & 97.6 & -\\
         AutoDIAL \cite{carlucci2017autodial} & & 97.96 & 97.51 & 89.12 & 10.78 \\
         SBADA-GAN \cite{russo17sbadagan} & & 97.6 & 95.0 & 76.1 & \textbf{61.1}\\
         GAM \cite{huang2018domain} & & 95.7$\pm$0.5 & 98.0$\pm$0.5 & 74.6$\pm$1.1 & - \\
         MECA \cite{morerio2017minimal} && - & - & 95.2 & - \\
         \textbf{DWT} & & \textbf{99.09}$\pm$0.09 & \textbf{98.79}$\pm$0.05 & \textbf{97.75}$\pm$0.10 & 28.92 $\pm$1.9 \\
         \hline
         Target Only & & 96.5 & 99.2 & 99.5 & 96.7\\
         \hline
         w/ augmentation\\
         SE \textsuperscript{a} \cite{french2018self} & & 88.14$\pm$0.34 & 92.35$\pm$8.61 & 93.33$\pm$5.88 & 33.87$\pm$4.02 \\
         
         SE \textsuperscript{b} \cite{french2018self} & &
         98.23$\pm$0.13 & \textbf{99.54}$\pm$0.04 & \textbf{99.26}$\pm$0.05 & \textbf{37.49}$\pm$2.44 \\
         
         SE \textsuperscript{\dag} \textsuperscript{b} \cite{french2018self} & &
         99.29$\pm$0.16 & 99.26$\pm$0.04 & 97.88$\pm$0.03 & 24.09$\pm$0.33 \\
         
         
         \textbf{DWT-MEC}\textsuperscript{b} & & 99.01$\pm$0.06 & 99.02$\pm$0.05 & 97.80$\pm$0.07 & 30.20$\pm$0.92\\
         
         
         \textbf{DWT-MEC (MT)}\textsuperscript{b} & & \textbf{99.30}$\pm$0.19 & 99.15$\pm$0.05 & 99.14$\pm$0.02 & 31.58$\pm$2.34\\
         
         \hline
         \hline
         
    \end{tabular}
    \vspace{2mm}
    \caption{Accuracy (\%) on the digits datasets: comparison with state of the art. \textsuperscript{a} indicates minimal usage of data augmentation and \textsuperscript{b} considers augmented source and target data. 
    \textsuperscript{\dag} indicates our implementation of SE \cite{french2018self}. }
    \label{tab:small_scale_exp}
\end{table*}

\vspace{-3mm}
\subsubsection{Ablation Study}
We first conduct a thorough analysis of our method assessing, in isolation, the impact of our two main contributions: (i) aligning source and target distributions by embedded DWT layers; and (ii) leveraging target data through our threshold-free MEC loss.

\begin{figure}
    \centering
    \includegraphics[width=1\columnwidth]{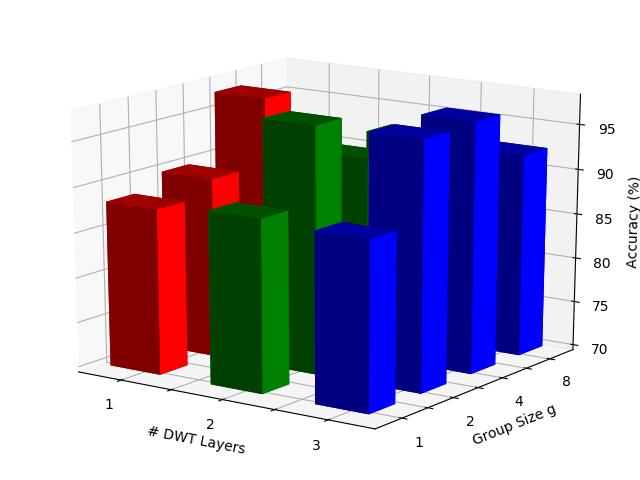}
    \caption{SVHN $\rightarrow$ MNIST experiment: accuracy at varying number of DWT layers and group size. Different colors are used to improve readability. 
    }
    \label{fig:wh_ablation}
\end{figure}

First, we consider the SVHN$\rightarrow$MNIST setting and we show the benefit of feature whitening over BN. We vary the number of whitening layers from 1 to 3 and simultaneously change the group size ($g$) 
from 1 to 8 (see Sec.~\ref{sec:implementation_details}). With group size equal to 1, DWT layers reduces to DA layers as proposed in \cite{carlucci2017autodial,li2016revisiting}. 
Our results are shown in Fig.~\ref{fig:wh_ablation} and from the figure it is clear that when $g=1$ the accuracy stays consistently below 90$\%$. This behaviour can be ascribed to the sub-optimal alignment of source and target data distributions achieved with previous BN-based DA layers. 
When the group size increases, the feature decorrelation  performed by the DWT layers comes into play and results into a significant improvement in terms of performance. 
The accuracy increases monotonically as the group size grows until the  value of $g=4$, then it start to decrease. 
This final drop is probably due to 
ill-conditioned covariance matrices. Indeed, a covariance matrix with size 8 $\times$ 8 is perhaps poorly estimated due to the lack of samples in a batch (Sec.~\ref{sec:implementation_details}). 
Importantly, Fig.~\ref{fig:wh_ablation} also shows that increasing the number of DWT layers has a positive impact on the accuracy. 
This is in contrast with \cite{huang2018decorrelated}, where feature decorrelation is used only in the first layer of the network. 


\begin{table}[h]
    \centering
    \setlength{\tabcolsep}{4pt}
    \begin{tabular}{lccccc}
        \hline
        \hline
         Method & \shortstack{Source\\Target} & \shortstack{MNIST\\USPS} & \shortstack{USPS\\MNIST}  & \shortstack{SVHN\\MNIST}  \\
         \hline
         SE (w/ CT) \cite{french2018self} && 99.29 & 99.26 & 97.88\\
         SE (w/o CT) \cite{french2018self} && 98.71 & 97.63 & 26.80 \\
         \hline
         \textbf{DWT-MEC (MT)} && \textbf{99.30} & \textbf{99.15} & \textbf{99.14} \\
         \hline
         \hline
    \end{tabular}
    \vspace{0.5mm}
    \caption{Accuracy (\%) on the digits datasets. Comparison between the consistency loss in SE method \cite{french2018self} (with and without CT) and our threshold-free MEC loss.}
    \label{tab:zl_ablation}
    \vspace{-0.4cm}
\end{table}

In Tab.~\ref{tab:zl_ablation} we evaluate the effectiveness of the proposed MEC loss and we compare our approach with the consistency based loss adopted  by French \etal \cite{french2018self}. We use Self-Ensembling (SE) \cite{french2018self} with and without confidence thresholding (CT) on the network predictions of the teacher network. To fairly compare our approach with SE we also consider a mean-teacher scheme in our framework.
We observe that SE have excellent performance when the CT is set to a very high value (0.936 as in \cite{french2018self}) but it performance drops when CT is set equal to 0, especially in the SVHN$\rightarrow$MNIST setting. This shows that the consistency loss in \cite{french2018self} may be harmful when the network is not confident on the target samples. Conversey, the proposed MEC loss leads to results which are on par to SE in the MNIST$\leftrightarrow$USPS settings and to higher accuracy in the SVHN$\rightarrow$MNIST setting. This clearly demonstrates that our proposed loss avoids the need of introducing the CT hyper-parameter and, at the same time, yields to better performance. It is important to remark that, in the case of UDA, tuning hyper-parameters is hard as target samples are unlabeled and cross-validation on source data is unreliable because of the domain shift problem \cite{morerio2017minimal}.

\begin{table*}[t]
        \centering
	    \setlength{\tabcolsep}{4pt}
        \begin{tabular}{l|ccccccccccccc|c}
            \hline
            \hline
            Method & \shortstack{ Source \\ Target} & \shortstack{Ar \\ Cl} & \shortstack{Ar\\Pr} & \shortstack{Ar\\Rw} & \shortstack{Cl\\Ar} & \shortstack{Cl\\Pr} & \shortstack{Cl\\Rw} & \shortstack{Pr\\Ar} & \shortstack{Pr\\Cl} & \shortstack{Pr\\Rw} & \shortstack{Rw\\Ar} & \shortstack{Rw\\Cl} & \shortstack{Rw\\Pr} & Avg\\
            \hline
            ResNet-50 \cite{he2016deep} && 34.9 & 50.0 & 58.0 & 37.4 & 41.9 & 46.2 & 38.5 &  31.2 & 60.4 & 53.9 & 41.2 &  59.9 & 46.1\\
            DAN	\cite{long2015learning} && 43.6 & 57.0 & 67.9 & 45.8 & 56.5 & 60.4 & 44.0 &  43.6 & 67.7 & 63.1 & 51.5 & 74.3 & 56.3\\
            DANN \cite{ganin2016domain} &&  45.6 & 59.3 & 70.1 & 47.0 & 58.5 & 60.9 & 46.1 & 43.7 &  68.5 & 63.2 & 51.8 & 76.8 & 57.6\\
            JAN \cite{long2016deep} && 45.9 &  61.2 & 68.9 & 50.4 & 59.7 & 61.0 &  45.8 & 43.4 & 70.3 & 63.9 & 52.4 & 76.8 & 58.3\\
            SE \cite{french2018self} && 48.8 & 61.8 & 72.8 & 54.1 & 63.2 & 65.1 & 50.6 & 49.2 &  72.3 & 66.1 & 55.9 & 78.7 & 61.5 \\
            CDAN-RM \cite{long2017conditional} && 49.2 & 64.8 & 72.9 & 53.8& 63.9 & 62.9 & 49.8 & 48.8 &  71.5 & 65.8 &  56.4 & 79.2 & 61.6\\
            CDAN-M \cite{long2017conditional} && \textbf{50.6} & 65.9 & 73.4 & 55.7 &  62.7 & 64.2 & 51.8 &  \textbf{49.1} &  74.5 & 68.2 & \textbf{56.9} & 80.7 & 62.8\\
            
            \textbf{DWT-MEC} && 50.3 & \textbf{72.1} & \textbf{77.0} & \textbf{59.6} & \textbf{69.3} & \textbf{70.2} & \textbf{58.3} & 48.1 & \textbf{77.3} & \textbf{69.3} & 53.6 & \textbf{82.0} & \textbf{65.6} \\
            \hline
            \hline
        \end{tabular}
        \vspace{2mm}
        \caption{Accuracy(\%) on Office-Home dataset with Resnet-50 as base network and comparison with the state-of-the-art methods.}
        \label{tab:resnet50}
    \end{table*}

\begin{table}[t]
    \centering
	\setlength{\tabcolsep}{5pt}
    \begin{tabular}{l|cccc}
        \hline
        \hline
         & \shortstack{Source\\Target} & \shortstack{CIFAR-10\\STL} & \shortstack{STL\\CIFAR-10}\\ 
         \hline
         \hline
         Source Only &  & 60.35 & 51.88\\
         \hline
         w/o augmentation\\
         DANN \cite{ganin2016domain} & & 66.12 & 56.91\\
         DRCN \cite{ghifary2016deep} & & 66.37 & 58.65\\
         AutoDIAL \cite{carlucci2017autodial} &  & 79.10 & 70.15\\
         \textbf{DWT}  & &\textbf{79.75}$\pm$0.25 & \textbf{71.18}$\pm$0.56\\
         \hline
         Target Only &  & 67.75 & 88.86 \\
         \hline
         w/ augmentation\\
         SE \textsuperscript{a} \cite{french2018self} &  & 77.53$\pm$0.11 & 71.65$\pm$0.67 \\
   
         SE \textsuperscript{b} \cite{french2018self} &  & 80.09$\pm$0.31 & 69.86$\pm$1.97\\
         
         
         \textbf{DWT-MEC}\textsuperscript{b} & & 80.39$\pm$0.31  & \textbf{72.52}$\pm$0.94 \\
         
         
          \textbf{DWT-MEC (MT)}\textsuperscript{b} & & \textbf{81.83}$\pm$0.14  & 71.31$\pm$0.22 \\
         
         \hline
         \hline

         \hline
    \end{tabular}
    \vspace{2mm}
    \caption{Accuracy (\%) on the CIFAR-10$\leftrightarrow$STL: comparison with state of the art. \textsuperscript{a} indicates minimal  data augmentation and \textsuperscript{b} considers augmented source and target data.
    }
    \label{tab:cifar_stl_exp}
        \vspace{-0.4cm}
\end{table}

\vspace{-1mm}
\subsubsection{Comparison with State-of-the-Art Methods}
\label{sec.SOTA}

In this section we present our results and compare with previous UDA methods. 
Tab.~\ref{tab:small_scale_exp} reports the results obtained on the digits datasets. We compare with several baselines: Correlation Alignment (CORAL) \cite{sun2016return}, Simultaneous Deep Transfer (MMD) \cite{tzeng2015simultaneous}, Domain-Adversarial Training of Neural Networks (DANN) \cite{ganin2016domain}, Domain separation networks \cite{bousmalis2016domain}, Coupled generative adversarial net-works (CoGAN) \cite{liu2016coupled}, Adversarial discriminative domain adaptation (ADDA) \cite{tzeng2017adversarial}, Deep reconstruction-classification networks (DRCN),  \cite{ghifary2016deep}, Asymmetric  tri-training \cite{saito2017asymmetric}, Associative domain adaptation (ADA) \cite{haeusser2017associative}, AutoDIAL \cite{carlucci2017autodial}, SBADA-GAN \cite{russo17sbadagan}, Domain transferthrough deep activation matching (GAM) \cite{huang2018domain}, Minimal-entropy correlation alignment (MECA) \cite{morerio2017minimal} and SE \cite{french2018self}. Note that the Virtual Adversarial
Domain Adaptation (VADA) \cite{shu2018dirt} use a different network, thus cannot be compared with the other methods (including ours) which are based on a different capacity network. For this reason, \cite{shu2018dirt}
is not reported in Tab.~\ref{tab:small_scale_exp}. 
 Results associated with each method are taken from the corresponding papers. 
 We re-implemented  SE 
 as the numbers reported in the original paper \cite{french2018self} refer to different network architectures. 

Tab.~\ref{tab:small_scale_exp} is split in two  sections,  separating those methods that exploit data augmentation from those which use only the original training data. Compared with no-data augmentation methods, our DWT  performs better than previous UDA methods in the three settings. Our method is less effective in the MNIST$\rightarrow$SVHN due to the strong domain shift between the two domains. In this setting, GAN-based methods \cite{russo17sbadagan} are more effective. 
Looking at methods which consider data augmentation, we compare our approach with SE \cite{french2018self}. To be consistent with other methods, we plug  the architectures described in \cite{ganin2014unsupervised} in SE. Comparing the proposed approach with our re-implementation of SE (SE$\textsuperscript{\dag}\textsuperscript{b}$) we observe that DWT-MEC (MT) is almost on par with SE in the MNIST$\leftrightarrow$USPS setting and better than SE in the SVHN$\rightarrow$MNIST. 
For the sake of completeness, we also report the performance of SE taken from the original paper \cite{french2018self}, considering SE with minimal augmentation (only gaussian blur) and SE with full augmentation. 

With the rapid progress of deep DA methods, the results in the digits datasets have saturated. This makes it difficult to gauge the merit of the proposed contributions. Therefore, we also consider the CIFAR10 $\leftrightarrow$ STL setting. 
Our results are reported in Tab.~\ref{tab:cifar_stl_exp}.
Similarly to the experiments in Tab.~\ref{tab:small_scale_exp}, we  separate those methods exploiting data augmentation from those not using target-sample perturbations. Tab.~\ref{tab:cifar_stl_exp} shows that our method (DWT), outperforms all previous baselines which also do not consider augmentation. Furthermore, by exploiting data perturbation and the proposed MEC loss our approach (with and without Mean-Teacher) reaches higher accuracy than SE.\footnote{In this case the accuracy values reported for SE are taken directly from the original paper as the underlying network architecture is the same.}

Finally, we also perform experiments on the large-scale Office-Home dataset and we compare with the baselines methods as reported by Long \textit{et al.} \cite{long2017conditional}. 
The results  reported in Tab.~\ref{tab:resnet50} show that our approach outperforms all the other  methods. On average, the proposed approach improves over Conditional Domain Adversarial Networks (CDAN) by 2.8$\%$ and it is also more accurate than SE. 



 
\section{Conclusions}
\label{sec:conclusion}
In this work we address UDA by proposing domain-specific feature whitening with DWT layers and the MEC loss. On the one hand,  whitening of intermediate features enables the alignment of the source and the target distributions at intermediate feature levels and increases the smoothness of the loss landscape. On the other hand, our MEC loss better exploits the target data. 
Both these components can be easily integrated in any standard CNN. Our experiments on standard benchmarks show state-of-the-art performance on digits categorization and object recognition tasks. As future work, we plan to extend our method to handle multiple source and target domains.
\vspace{-3mm}
\subsubsection*{Acknowledgments}
This work was carried out under the \enquote{Vision and Learning joint Laboratory} between FBK and UNITN. We thank the NVIDIA Corporation for the donation of the GPUs used in this project. This project has received funding from: i) the European Research Council (ERC) (Grant agreement No.788793-BACKUP); and ii) project DIGIMAP, funded under grant \#860375 by the Austrian Research Promotion Agency (FFG).

{\small
\bibliographystyle{ieee}
\bibliography{egbib}
}

\end{document}